\documentclass[10pt,twocolumn,letterpaper]{article}

\usepackage{cvpr}              

\usepackage{graphicx}
\usepackage{amsfonts}
\usepackage{amsmath}
\DeclareMathOperator*{\argmax}{arg\,max}

\usepackage{amssymb}
\usepackage{bbm}
\usepackage{booktabs}
\usepackage[noend, ruled, lined, linesnumbered]{algorithm2e}

\SetCommentSty{commfont}
\usepackage{enumitem}
\setlist[itemize]{noitemsep, topsep=0pt}
\def\algo{\textbf{\texttt{MaskMatch}}\xspace}

\usepackage[pagebackref,breaklinks,colorlinks]{hyperref}
\usepackage{multirow}
\usepackage{makecell}
\usepackage[capitalize]{cleveref}
\crefname{section}{Sec.}{Secs.}
\Crefname{section}{Section}{Sections}
\Crefname{table}{Table}{Tables}
\crefname{table}{Tab.}{Tabs.}

\def\cvprPaperID{2347} 
\def\confName{CVPR}
\def\confYear{2024}

\begin{document}

\title{MaskMatch: Boosting Semi-Supervised Learning Through Mask Autoencoder-Driven Feature Learning
}

\author{Wenjin Zhang$^{1}$, Keyi Li$^{1}$, Sen Yang$^{2}$, Chenyang Gao$^{1}$, Wanzhao Yang$^{1}$, Sifan Yuan$^{1}$, Ivan Marsic$^{1}$\\
$^{1}$Rutgers University. $^{2}$Waymo.\\
\{wz315, kl734, sy358, cg694, wy209, sy609\}@scarletmail.rutgers.edu, marsic@rutgers.edu\\
}

\maketitle

\begin{abstract} 
Conventional methods in semi-supervised learning (SSL) often face challenges related to limited data utilization, mainly due to their reliance on threshold-based techniques for selecting high-confidence unlabeled data during training. Various efforts (e.g., FreeMatch) have been made to enhance data utilization by tweaking the thresholds, yet none have managed to use 100\% of the available data. To overcome this limitation and improve SSL performance, we introduce \algo, a novel algorithm that fully utilizes unlabeled data to boost semi-supervised learning. \algo integrates a self-supervised learning strategy, i.e., Masked Autoencoder (MAE), that uses all available data to enforce the visual representation learning. This enables the SSL algorithm to leverage all available data, including samples typically filtered out by traditional methods. In addition, we propose a synthetic data training approach to further increase data utilization and improve generalization. These innovations lead \algo to achieve state-of-the-art results on challenging datasets. For instance, on CIFAR-100 with 2 labels per class, STL-10 with 4 labels per class, and Euro-SAT with 2 labels per class, \algo achieves low error rates of 18.71\%, 9.47\%, and 3.07\%, respectively. The code will be made publicly available.




\end{abstract}
\section{Introduction} 

Semi-supervised learning (SSL) has been the focus of thorough investigation for an extended period, primarily due to its ability to leverage large quantities of unlabeled data~\cite{laine2017temporal, mean_teacher_2017, miyato2018VAT, xie2020UDA, Pseudolabel, fixmatch, wang2023freematch, zhang2022flexmatch}. This becomes especially beneficial in scenarios where the availability of labeled data is limited. Among the existing SSL techniques, 
pseudo-labeling~\cite{Pseudolabel, xie2020UDA, fixmatch, wang2023freematch, zhang2022flexmatch} is a widely used paradigm. The core idea is that if the model demonstrates confidence in its predictions for unlabeled data, it should produce similar predictions or pseudo labels for the same unlabeled data, even when subjected to various perturbations. Typically, confident data samples are identified using threshold-based methodologies~\cite{Pseudolabel,fixmatch, zhang2022flexmatch, wang2023freematch}. For instance, given an unlabeled sample, the current model predicts its confidence. This sample is employed in the later training process only if its confidence exceeds a predefined threshold.


\begin{figure}
    \centering
    \includegraphics[width=1\linewidth]{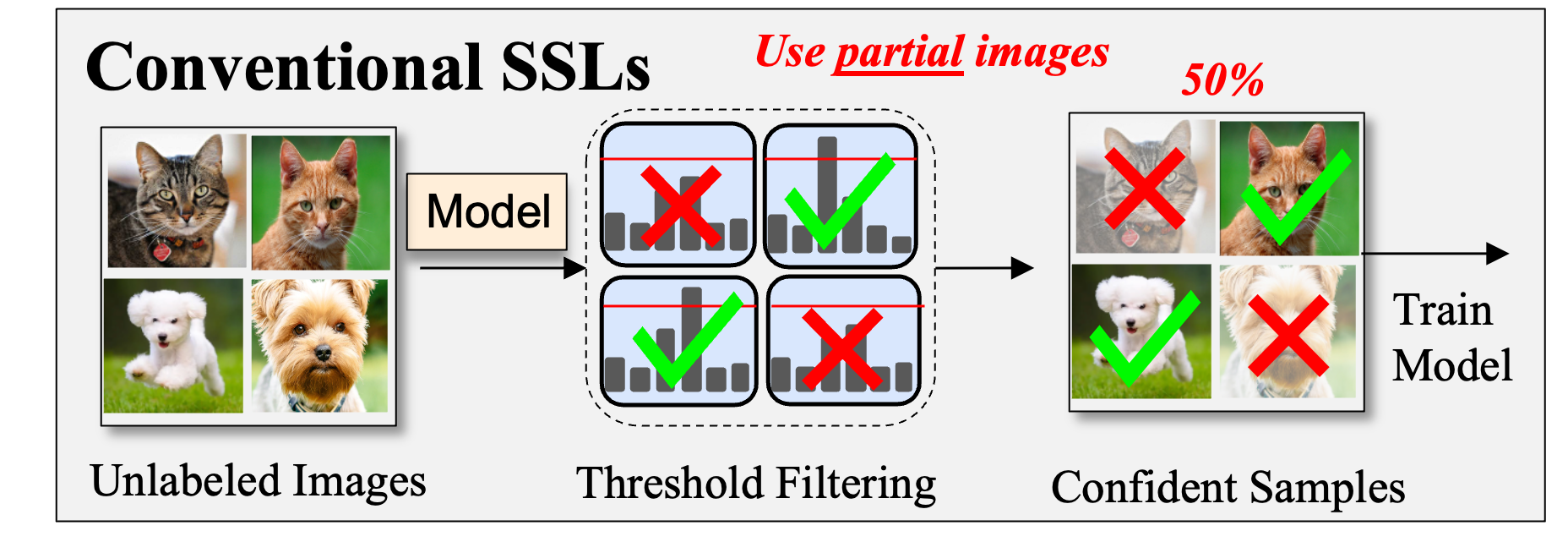}
    \vspace{-7mm}
    \caption{Conventional SSL algorithms with confidence-based thresholding only make use of a fraction of unlabeled images, as they filter data based on predefined thresholds.}
    \label{fig:method_compare}
    \vspace{-5mm}
\end{figure}

\begin{figure}
    \centering
    \includegraphics[width=0.85\linewidth]{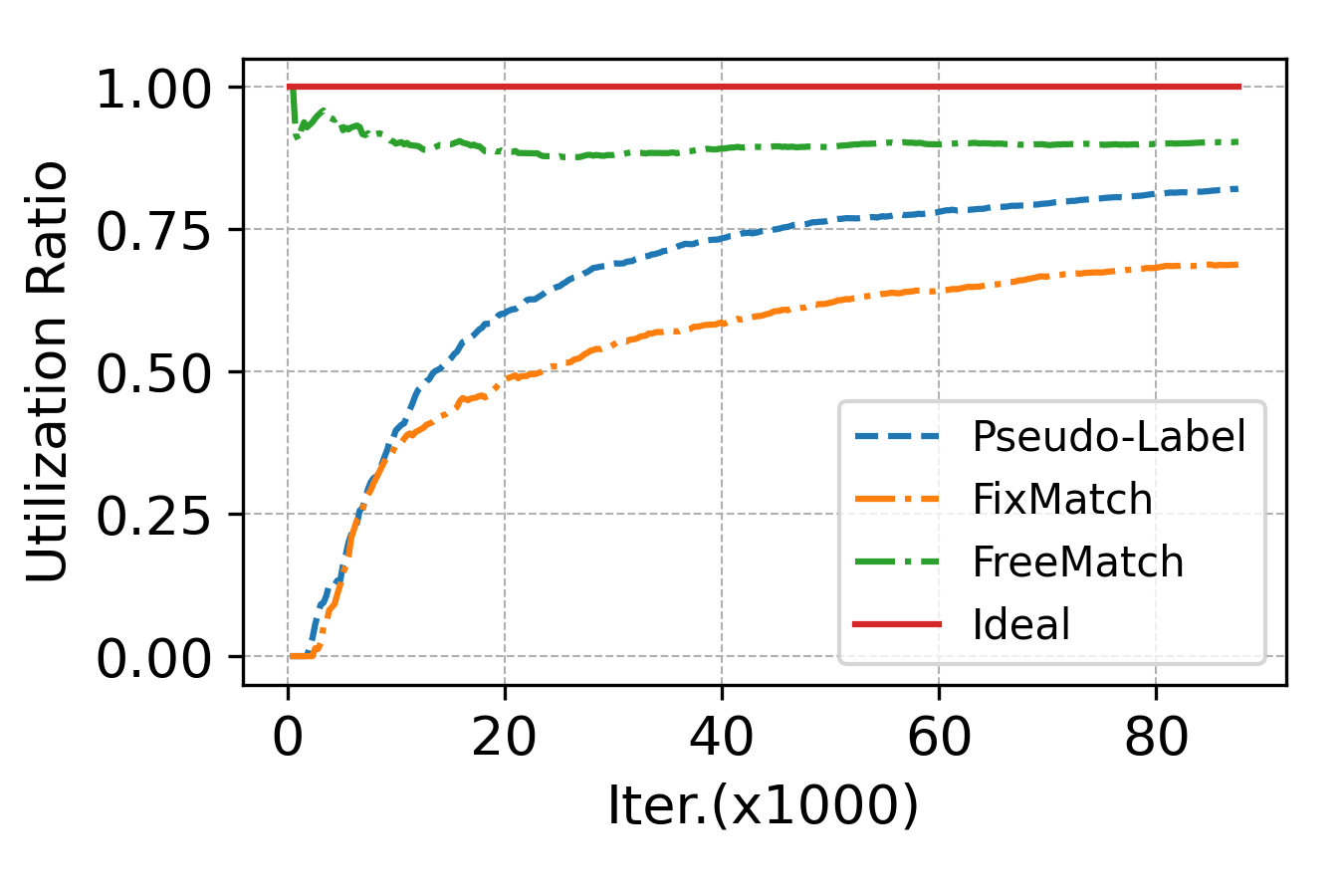}
    \vspace{-5mm}
    \caption{Unlabeled data utilization ratio of popular SSL algorithms on CIFAR-100 dataset with 2 labels per class}
    \label{fig:accum_utilization}
    \vspace{-5mm}
\end{figure}

One potential drawback of threshold-based methods is their dependency on a threshold for calculating the unsupervised loss. 
While this approach ensures that only high-quality unlabeled data influences model training, it neglects a substantial part of other unlabeled data (Figure~\ref{fig:method_compare}). This limitation confines the use of unlabeled data to instances with prediction confidence exceeding the specified threshold, potentially biasing the model and impacting generalization.
Despite attempts to introduce adaptive thresholds in recent studies~\cite{wang2023freematch, zhang2022flexmatch}, these methods still fall short of fully utilizing unlabeled data. When applying widely-used SSL algorithms on the CIFAR-100 dataset with 200 labeled and 50,000 unlabeled samples, we observe that fixed threshold-based approaches (e.g., PseudoLabel~\cite{Pseudolabel} and FixMatch~\cite{fixmatch}) only leverages a fraction of the available unlabeled data, leaving the uncertain unlabeled samples untouched (Figure~\ref{fig:accum_utilization}). 
The adaptive threshold approach, FreeMatch~\cite{wang2023freematch}, briefly reaches 100\% utilization with an initial low threshold (i.e., $\frac{1}{\#Class}$) in the early stage. It then gradually decreases to around 88\% utilization. A concern arises regarding the correctness of pseudo-labels derived from a very low threshold, which may misguide model training. 
\textbf{\textit{We hypothesize that the critical samples for model training are included in these uncertain samples.}} The above analysis motivates us to address the threshold limitation and utilize all unlabeled data to enhance semi-supervised training. 

We propose \algo, an SSL algorithm that can boost SSL performance by training the model with all unlabeled data, including the uncertain ones (Figure~\ref{fig:overall_approach}). The main challenge of utilizing uncertain samples is the lack of reliable pseudo-labels, which means there is no pseudo-supervised target. A key question explored here is: can we train the model from a different perspective beyond pseudo-labels? Inspired by Masked Autoencoder (MAE)~\cite{he2021masked} in the self-supervised learning area, we introduce an MAE-based reconstruction task into SSL to jointly train the model with all unlabeled data. In addition, we propose synthetic data training (SDT) to further increase data utilization and enhance the ability to generalize. The key contributions of this paper are summarized as follows:
\begin{itemize}[noitemsep,topsep=0pt]
    \item We propose \algo, which fully utilizes unlabeled data to improve SSL performance.   
    \item We develop a synthetic data training approach to further increase data utilization and improve generalization capabilities.
    \item We evaluate \algo across a broad set of benchmarks and show that \algo can boost SSL and outperform the existing SSL algorithms.
\end{itemize}

\section{Related Work}
\subsection{Semi-supervised Learning}
\label{sec:SSL}
Semi-supervised learning has been extensively researched over the last decades~\cite{ssl_Chapelle_2009, ssl_zhu_2008}. 
Recent work in this domain can be broadly categorized into two main types: consistency-based~\cite{laine2017temporal, mean_teacher_2017, miyato2018VAT, xie2020UDA} and pseudo-labeling~\cite{Pseudolabel, fixmatch, Pham_2021_CVPR}. Consistency-based approaches typically introduce perturbations either to the input or to the model itself and then impose constraints to ensure the coherence of extracted features or probability outputs. For instance, the $\Pi$-model~\cite{laine2017temporal} introduced noise to the model weights through dropout to construct two outputs for subsequent consistency regularization. On the other hand, pseudo-labeling leveraged model predictions as hard pseudo-labels to guide the learning process on unlabeled data and this concept has gained prominence in SSL recently~\cite{Pseudolabel, fixmatch, Pham_2021_CVPR, cai2021exponential, xie2020selftraining}. Depending on the timing of generating pseudo-labels, this method can be further divided into two categories: offline pseudo-labeling\cite{Pseudolabel, xie2020selftraining} and online pseudo-labeling~\cite{fixmatch, cai2021exponential}. In offline pseudo-labeling, 
pseudo-labels for unlabeled data are generated before the actual training begins and remain fixed throughout the training process. Conversely, online pseudo-labeling dynamically generates pseudo-labels for unlabeled data during the training process. Our approach uses online pseudo-labeling, enhanced with self-supervised training, to improve the performance by maximizing the utilization of unlabeled data.

\subsection{Self-supervised Learning} 
Self-supervised learning has attracted considerable attention in the field of computer vision. 
It is intricately tied to pretext tasks~\cite{Doersch_2015,Wang_2015, Noroozi_2016,Zhang_2016, Pathak_2017,gidaris2018unsupervised}. Their objective is to encourage models to capture meaningful data representations without relying on labeled examples. Two pre-training approaches (contrastive learning and masked image modeling) are popular. Contrastive learning~\cite{Becker_1992,Hadsell} revolves around modeling both the similarity and dissimilarity (or only similarity, as seen in~\cite{grill2020bootstrap, Chen_2021}) between two or more views. Notably, methods based on contrastive learning heavily rely on data augmentation~\cite{pmlr-v119-chen20j, grill2020bootstrap, Chen_2021}. Conversely, masked image modeling involves learning representations from images that are deliberately corrupted by masking image patches. For instance, Context Encoders in~\cite{Pathak_2016} trained a network to generate missing image patches based on their surrounding patches. Building on this concept, subsequent studies explored pre-training ViTs with Masked Autoencoders~\cite{pmlr-v119-chen20s, Xie_2022, he2021masked, Wei_2022, bao2022beit}. These approaches introduce mask-noise to images and predict missing input values at the pixel or patch levels. Among these, the most widely adopted method is Masked Autoencoder (MAE)\cite{he2021masked} because of its improving performance among various downstream tasks (e.g., image classification, segmentation, and object detection). Our work incorporates the MAE reconstruction loss as a constraint into semi-supervised learning. The reconstruction loss from all images (including unconfident ones) enhances visual representation learning and, consequently, improves the performance of semi-supervised learning.

\section{Preliminaries}

Following the semi-supervised learning paradigm~\cite{fixmatch, wang2023freematch, chen2023softmatch}, the model is jointly trained with a supervised loss on labeled data and an unsupervised loss on unlabeled data. 
We define the framework of SSL within the context of a $C$-class classification scenario. Let $\mathcal{X}=\left\{ \left( x_i, y_i\right) : i \in \left[ 1,B_l \right]\right\}$ and $\mathcal{U}=\left\{ u_i : i \in \left[ 1,B_{u} \right]\right\} $ represent a batch of labeled data and a batch of unlabeled data, where $B_l$ and $B_{u}$ are their batch sizes, $x_i$ and $y_i$ are labeled images and their one-hot labels, and $u_i$ represents the unlabeled images. The supervised loss on labeled data is:
\begin{equation}
  \mathcal{L}_{s} = \frac{1}{B_l}\sum_{i=1}^{B_l} \mathcal{H} \left( y_i, \mathcal{F}_{\theta} \left( y| \alpha (x_i) \right) \right),
  \label{eq:supervised_loss}
\end{equation}
where $\mathcal{H}$ is standard cross-entropy loss, $\alpha \left( \cdot \right)$ represents weakly augmentation functions, e.g., random crop and random flip, and $\mathcal{F}_{\theta} \left( \cdot \right)$ is the predicted probability over $C$-class from the model.


For the unlabeled data, the in-training model generates pseudo-labels for unlabeled data. Then, adaptive class-specific threshold~\cite{wang2023freematch} is used to avoid potentially wrong pseudo-labels misleading the model.
Let $\tau_t \in R^C$ represent the adaptive threshold at training iteration index $t$. Thus, the loss on unlabeled data is represented by:

\begin{equation}
    \label{eq:loss_unlabled}
    \begin{aligned}
        & \mathcal{L}_{u} = \frac{1}{B_u}\sum_{i=1}^{B_u} \mathcal{M}[i] \cdot \mathcal{H}\left( \hat{p}_i, P_i \right), \\
        & \mathcal{M} = \left\{ \mathbbm{1} \left( max(p_i) > \tau_t[\argmax{p_i}] \right): i \in [1, B_u]\right\},
    \end{aligned}
\end{equation}
where $p_i = \mathcal{F}_{\theta}\left( y| \alpha (u_i) \right)$ and $P_i = \mathcal{F}_{\theta}\left( y| \mathcal{A} (u_i) \right)$ represent the model output to the weakly and strongly augmented images, respectively. $\hat{p}_i$ represents the hard one-hot label obtained from $p_i$. $\mathcal{A}(\cdot)$ represents the strongly augmentation function, i.e., RandAugment ~\cite{cubuk2019randaugment}. $\mathcal{M}[i]$ is a pseudo-label selection mask, which is the output of the indicator function, $\mathbbm{1} \left( \cdot > \tau_t[\argmax{p_i}] \right)$, when applying the class-specific threshold $\tau_t \in R^C$. The details of class-specific threshold are discussed later in Section~\ref{sec:undateSAT}.


\begin{figure*}
    \centering
    \includegraphics[width=1\linewidth]{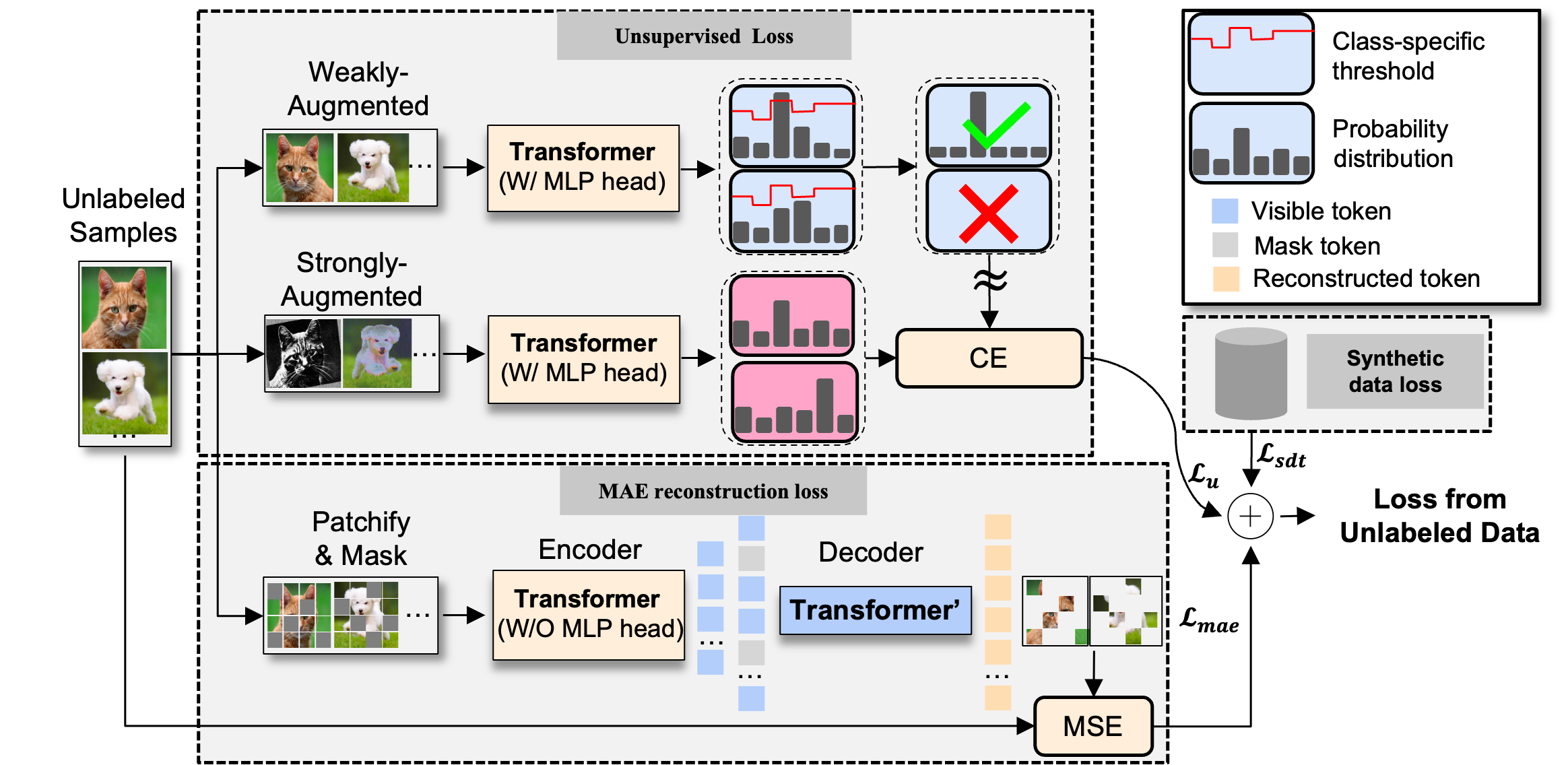}
    \caption{Diagram of \algo. An in-training model (i.e., Transformer) is trained with three loss terms from unlabeled images. First, the\textbf{
    unsupervised loss} is defined as the divergence between class probabilities generated from strongly augmented images and the corresponding predictions aligned with one-hot pseudo-labels derived from weakly augmented confident images. Second, \textbf{MAE reconstruction loss} is computed from all images. We patchify and randomly mask the images. The in-training model without an MLP classifier head (i.e., encoder) and decoder are trained by reconstructing these masked patches. The decoder is an auxiliary transformer for assisting encoder training only. Lastly, \textbf{synthetic data loss} is calculated by training the model on a synthetic dataset, a mixture of unlabeled and labeled images.}
    \label{fig:overall_approach}
    \vspace{-2mm}
\end{figure*}


\section{\algo}
\textbf{Overall framework.} Figure~\ref{fig:overall_approach} shows how \algo uses all unlabeled data to train the model. Given one batch of training data, the computation procedure of total training loss is described in Algorithm~\ref{alg:maskedmatch}. Firstly, we introduce an MAE-based reconstruction task to train the model using all image data. A reconstruction loss, $\mathcal{L}_{mae}$ represents the reconstruction target. Secondly, we propose synthetic data training (SDT) to increase data usage further and improve generalization.
The synthetic loss is denoted by $\mathcal{L}_{sdt}$. Lastly, we modify the self-adaptive threshold method \cite{wang2023freematch} for better performance. Overall, the model training minimization objective is:
\begin{equation}
    \mathcal{L}_{total} =  \mathcal{L}_{s} + \lambda_{u} \mathcal{L}_{u} + \lambda_{mae} \mathcal{L}_{mae} + \lambda_{sdt} \mathcal{L}_{sdt},
\label{equ:total_loss}
\end{equation}
where $\lambda_{u}$, $ \lambda_{mae}$ and $ \lambda_{sdt}$ are the loss coefficients for $\mathcal{L}_{u}$, $\mathcal{L}_{mae}$ and $\mathcal{L}_{sdt}$, respectively. Loss coefficients control the training bias to each loss term. The details for our loss coefficient settings are in Section~\ref{sec:setup}. Here are some of our rules for adjusting the coefficients:
\begin{itemize}
   \item To be consistent with conventional SSLs implementations in~\cite{wang2022usb}, we set $\lambda_{u} = 1 $.
   \item To avoid synthetic data loss dominating training, let $\lambda_{sdt} \approx \frac{\overline{\mathcal{L}}_{s} + \lambda_{u} \overline{\mathcal{L}}_{u}}{\overline{\mathcal{L}}_{sdt}}$, where $\overline{\mathcal{L}}_{s}$, $\overline{\mathcal{L}}_{u}$, and $\overline{\mathcal{L}}_{sdt}$ are their loss magnitude when they converge.
   \item Given that our target is an image classification task rather than a reconstruction task, we want $\lambda_{mae} < \frac{\overline{\mathcal{L}}_{s} + \lambda_{u} \overline{\mathcal{L}}_{u}}{\overline{\mathcal{L}}_{mae}}$, where  $\overline{\mathcal{L}}_{mae}$ is MAE loss magnitude when it converges.
\end{itemize}

\subsection{MAE Reconstruction Loss}

MAE reconstruction loss aims to improve the in-training model’s ability to feature extraction by learning from all images. This process involves random masking and subsequent reconstruction through an encoder (the in-training model without an MLP classifier head) and a decoder (an auxiliary transformer). The training process with MAE reconstruction does not require labels. Consequently, \algo has the ability to learn from all images, including those that would typically be discarded by confident thresholds during training.

\textbf{Input of MAE loss.} In each iteration of training, we collect both labeled images and unlabeled images using only weak augmentation as the input $\mathcal{D} = \mathcal{X} \cup \mathcal{U} $ to calculate MAE loss at Line \ref{alg:maeinput} of Algorithm \ref{alg:maskedmatch}. Although labeled images are used to train the model in a supervised manner, these images may also improve the model's feature extraction capability in an unsupervised manner. 

\textbf{Patchify and mask.} Following standard ViT~\cite{vit2021}, each image is divided into a sequence of non-overlapping patches, (i.e., tokens), $\left\{ I(k) | k \in \left[1, N_k \right] \right\}$ where $N_k$ is the number of tokens. 
Before the training process, we employ a configurable masking ratio to randomly mask patches ($\left\{ I(k) | k \in \Omega \right\}$, where $\Omega$ denotes the selected masking indices). The remaining unmasked tokens are then denoted by $\left\{ I(k) | k \notin \Omega \right\}$. 

\textbf{Compute reconstruction loss.} The reconstruction target is to predict the masked tokens using the unmasked tokens, as shown in the bottom part of Figure \ref{fig:overall_approach}.
In detail, the in-training model takes only visible tokens (i.e., unmasked tokens) as input and embeds them like a standard ViT through sequence procedures (e.g., linear projection, adding position embedding, and applying Transformer blocks). Since the MLP classifier head is not applied, the output of the encoder is the embedding of visible tokens, denoted by $\left\{ E(k) | k \notin \Omega \right\}$. Different from the encoder, the decoder input is a full set of token embeddings: $\left\{ E(k) | k \notin \Omega \right\} \cup \left\{ E_{mask} | k \in \Omega \right\}$, where $E_{mask}$ is a shared mask token embedding (i.e., a learnable vector) added to the position of masked tokens as placeholders. The decoder outputs the reconstructed images, and the reconstructed masked patches are denoted by $\left\{ \hat{I}(k) | k \in \Omega \right\}$.
The primary purpose of this decoder is to facilitate the training of the encoder. Once the training is complete, the decoder is discarded. Thus, the design of the decoder is independent of the in-training model. In our ablation study, we study the decoder design (i.e., the number of transformer blocks). 
Finally, we use mean square error as the training objective, and the loss is shown as follows:
\begin{equation}
\mathcal{L}_{mae} = \frac{1}{\Omega}\sum_{k \in \Omega}^{}\left| Norm(I(k)) - \hat{I}(k) ) \right|^2,
\end{equation}
where $Norm(I(k)) = \frac{I(k) - mean(I(k))}{ std(I(k))}$ is an optional operation, normalizing pixel values of a given patch using mean and standard deviation. 

\subsection{Synthetic Data Training}
Synthetic data training aims to increase unlabeled data utilization and avoid overfitting. This idea is based on MixUp augmentation, which is a popular data augmentation approach for SSL methods~\cite{NEURIPS2019_mixMatch, berthelot2020remixmatch, cai2022semisupervisedvit}. 
A recent study~\cite{liu2023overtraining_mixup} reveals a weakness of MixUp augmentation. They discover that over-training with MixUp may hurt generalization, especially when data size is limited. This issue arises because MixUp training initially learns ``clean patterns" but eventually leads to overfitting on the ``noisy patterns". This study suggests switching from MixUp training to vanilla empirical risk minimization (i.e., no MixUp augmentation) at an appropriate time. However, determining the time of overfitting is still challenging. To address this challenge, we propose SDT, which jointly utilizes MixUp and no MixUp augmentation to train the model. 
Specifically, we utilize MixUp to create a new synthetic dataset, which is then fed into the model to produce a separate loss term, $\mathcal{L}_{sdt}$. Simultaneously, the loss derived from the original unlabeled images and pseudo-labels,
$\mathcal{L}_{u}$, is also in the overall training objective.

In detail, we first collect all samples passing the class-specific threshold into a clean set, $\mathcal{\widehat{U}}_{clean}= \left\{(u_i, \hat{p}_i): \mathcal{M}[i] = 1 \And i \in B_u \right\}$, where $\hat{p}_i$ is one-hot pseudo-label of $u_i$. We name the remaining samples as noisy set, $\mathcal{\widehat{U}}_{noisy} = \mathcal{U} - \mathcal{\widehat{U}}_{clean} $.
The synthetic data is generated from the clean set and labeled data.
Specifically, we merge the clean set and labeled data into a set, $S$. 
The synthetic dataset, $S^{'} = \left\{ (x^{'}_{i}, \hat{p}^{'}_{i}) : i \in [1, len(S)] \right\}$ is obtained through:
\begin{equation}
    \begin{aligned}
        x^{'}_{i} & = \Lambda x_{i} +  (1-\Lambda) x_{j}, \\ 
        \hat{p}^{'}_{i} & = \Lambda \hat{p}_{i} +  (1-\Lambda) \hat{p}_{j}, \\
    \end{aligned}
    \label{eq:mixup}
\end{equation}
where $(x_{i}, \hat{p}_{i}), (x_{j}, \hat{p}_{j}) \in S $, $i \neq j$, $j \in [1, len(S)] $, $ \Lambda = max(\Lambda^{'}, 1-\Lambda^{'}), \Lambda^{'} \sim Beta(0.5, 0.5) \in (0, 1)$. Let $\Lambda > 0.5 $ to ensure $x^{'}_{i}$ is close to $x_{i}$. Once we obtain $S^{'}$, we compute synthetic loss like computing supervised loss:
\begin{equation}
\mathcal{L}_{sdt} = \frac{1}{len(S^{'})}\sum_{i=1}^{len(S^{'})} \mathcal{H} \left( \hat{p}^{'}_{i}, \mathcal{F}_{\theta} \left( y| x^{'}_{i} \right) \right).
\label{eq:loss_sdt}
\end{equation}

\textbf{Difference to MixUp augmentation.} Our work is different from MixUp augmentation in~\cite{NEURIPS2019_mixMatch} and Probabilistic Pseudo Mixup augmentation in~\cite{cai2022semisupervisedvit}. While these methods only utilize augmented data for training, our approach incorporates original unlabeled samples and synthetic data. Moreover, in contrast to the Probabilistic Pseudo Mixup augmentation, which mixes clean and noisy sets according to confidence probability, our synthetic data is generated by mixing the clean set with labeled images.


\subsection{Modified Class-specific Threshold}
\label{sec:undateSAT}
\textbf{Revisiting class-specific threshold.}
FreeMatch~\cite{wang2023freematch} suggests the threshold reflects the learning status and should be updated according to the model’s confidence on unlabeled data. The threshold of specific class $\tau_t(c) $ is computed from global $\tau_t $ and local thresholds $\nu_t(c) $:
\begin{equation}
    \tau_t(c) = \frac{\nu_t(c)}{max\left\{ \nu_t(c): c \in [1, C] \right\}} \cdot \tau_t. 
    \label{equ:threshold}
    \vspace{-2mm}
\end{equation}
The global threshold is updated with training iteration:
\begin{equation}
    \tau_t = 
        \begin{cases}
        \frac{1}{C}, & t=0,\\
        \lambda \tau_{t-1} + (1 - \lambda) \frac{1}{B_u}\sum_{i=1}^{B_u}max(p_i)), & t>0,\\
        \end{cases}
\end{equation}
where $\lambda \in (0, 1)$ is the momentum decay of the exponential moving average (EMA). The local thresholds are adjusted as follows:
\begin{equation}
    \nu_t(c) = 
        \begin{cases}
        \frac{1}{C}, & t=0,\\
        \lambda \nu_{t-1}(c) + (1 - \lambda) \frac{1}{B_u}\sum_{i=1}^{B_u}p_i(c)), & t>0,\\
        \end{cases}
\end{equation}
From Eq. \ref{equ:threshold}, when $t=0$, the threshold is initially set as $\frac{1}{C}$. Such a low threshold in the early stage may lead to wrong pseudo-labels. 

\textbf{Modifying initial threshold.} 
\algo requires more stable pseudo-labels due to SDT. SDT mixes unlabeled samples with corresponding confident pseudo-labels and labeled images to generate synthetic training data. So, wrong pseudo-labels can reduce model performance because of wrong unsupervised and synthetic loss. To mitigate the impact of incorrect pseudo-labels at the early stage, we set $\nu_0(c)=1$ for $ c \in [1, C]$ and $\tau_0 = 1$. In the early stage, the model is trained on supervised loss and reconstruction loss. Then, the threshold will decrease gradually from 1 and increase in response to changes in the model confidences.

\begin{algorithm}
\small 
\SetKwInOut{Input}{Input}\SetKwInOut{Output}{Output}\SetKwInput{Config}{Config}
\Input{Labeled data: $\mathcal{X}=\left\{ \left( x_i, y_i\right) : i \in \left[ 1,B_l \right]\right\}$,
unlabeled data: $\mathcal{U}=\left\{ u_i : i \in \left[ 1,B_{u} \right]\right\}$.}
\Config{Beta distribution for MixUp: $Beta(0.5, 0.5)$, loss coefficients: $\lambda_{u}$, $ \lambda_{mae}$ and $ \lambda_{syn}$, number of classes: $C$, and masking ratio: $r_{mask}$.
}
\Output{Total training loss: $\mathcal{L}_{total}$.}
\BlankLine

Compute $\mathcal{L}_{s}$ from $\mathcal{X}$\tcp*{see Eq.~\ref{eq:supervised_loss}.}


Compute $\mathcal{M}$ and $\mathcal{L}_{u}$ from $\mathcal{U}$ \tcp*{see Eq.~\ref{eq:loss_unlabled}.}



 Compute $\mathcal{L}_{mae}$ from $Concat(\mathcal{X},\mathcal{U})$ with $r_{mask}$\; \label{alg:maeinput}


Select clean set $\mathcal{\widehat{U}}_{clean}$\; 
\label{alg:syn_data}

$S = Concat(\mathcal{X},\mathcal{\widehat{U}}_{clean})$\; 



Generate synthetic dataset $S^{'}$ from $S$ \tcp*{see Eq.~\ref{eq:mixup}.}

Compute $\mathcal{L}_{sdt}$ from $S^{'}$\tcp*{see Eq.~\ref{eq:loss_sdt}.}

$\mathcal{L}_{total} = \mathcal{L}_{s} + \lambda_{u} \mathcal{L}_{u} + \lambda_{mae} \mathcal{L}_{mae} + \lambda_{sdt} \mathcal{L}_{sdt}$

\Return $\mathcal{L}_{total}$
\caption{Training Loss of \algo}
\label{alg:maskedmatch}
\end{algorithm}

\begin{table}[ht]
\vspace{-2mm}
\centering
\caption{Details of Datasets}
\vspace{-3mm}
\label{tab:dataset}
\resizebox{\columnwidth}{11mm}{
\begin{tabular}{ccccc}
\hline
 
 \multirow{2}{*}{\textbf{Dataset}} & \multirow{2}{*}{\textbf{\#Classes}}& \multirow{2}{*}{\textbf{\#Labels per class}} & \textbf{\#Train images} & \multirow{2}{*}{\textbf{\#Test images}} \\ \cline{4-4}
 & & &\#Labeled/\#Unlabeled &\\
\hline
 CIFAR-100 & 100 & 2 , 4 & 50,000 & 10,000  \\
 STL-10 & 10 & 4 , 10 & 5,000 / 100,000 & 8,000  \\
 Euro-SAT & 10 & 2 , 4 & 16,200 & 5,400  \\
 TissueMNIST & 8 & 10 , 50 & 165,466 & 47,280  \\
 Semi-Aves & 200 & 15-53 & 5,959 / 26,640 & 4,000  \\
\hline
\end{tabular}}
\vspace{-5mm}
\end{table}
\section{Experiment}
\subsection{Setup}
\label{sec:setup}
\textbf{Datasets.}
We extensively evaluate the performance of \algo on current challenging SSL benchmarks~\cite{wang2022usb}, including CIFAR-100~\cite{cifar100}, STL-10~\cite{stl10}, Euro-SAT~\cite{Helber_2018_EuroSAT, Helber_2019_EuroSAT}, TissueMNIST~\cite{Yang_2021_MedMNIST, Yang_2023_MedMNISTV2} and Semi-Aves~\cite{su2021semisupervised}. Following~\cite{wang2022usb}, previous popular SSL datasets CIFAR-10~\cite{cifar100} and SVHN~\cite{37648SVHN} are excluded from evaluation consideration since state-of-the-art SSL algorithms~\cite{xie2020UDA, fixmatch, xu2021dash} on these datasets have demonstrated comparable performance to fully-supervised training. The details of the datasets are shown in Table~\ref{tab:dataset}. For each dataset, two settings of \#labels per class are selected (for example, 2 and 4 labels per class for CIFAR-100), except for Semi-Aves, which has a long-tailed data distribution. 
Labeled data are randomly sampled from the training data for each dataset except STL-10 and Semi-Aves. This is because the split of labeled and unlabeled data is predetermined (e.g., 5,959 labeled images and 26,640 unlabeled images in Semi-Aves). We used the default test set for evaluation~\cite{wang2022usb}. 


\textbf{Baseline SSL methods.} We extensively compare our method with popular SSL algorithms including Pseudo Labeling~\cite{Pseudolabel}, Mean Teacher~\cite{mean_teacher_2017}, $\Pi$ Model~\cite{rasmus2015semisupervised_pimodel}, VAT~\cite{miyato2018VAT}, MixMatch~\cite{NEURIPS2019_mixMatch}, ReMixMatch~\cite{berthelot2020remixmatch}, UDA~\cite{xie2020UDA}, FixMatch~\cite{fixmatch}, Dash~\cite{xu2021dash}, CoMatch~\cite{li2021comatch}, CRMatch~\cite{fan2021crmatch}, FlexMatch~\cite{zhang2022flexmatch}, AdaMatch~\cite{berthelot2022adamatch}, SimMatch~\cite{zheng2022simmatch}, FreeMatch~\cite{wang2023freematch}, SoftMatch~\cite{chen2023softmatch} and DefixMatch~\cite{schmutz2023dont}. As discussed in Section~\ref{sec:SSL}, these SSL algorithms represent various adaptations of consistency regularization and pseudo-labeling strategies, and they achieve state-of-the-art performance on common benchmarks. We additionally incorporate one benchmark: fully supervised. This method employs all images and their labels (not limited to given labeled images) for training. Thus, the fully supervised method represents the upper bound of performance that SSLs can achieve.

\textbf{Implementation.} For a fair comparison, we train and evaluate all methods using the implementation from a unified codebase, USB~\cite{wang2022usb}. All methods are evaluated using the same network architectures and hyper-parameters for network training suggested by USB. In detail, we use ViT-Tiny with a patch size of 2 and image size of 32 for TissueMNIST;
ViT-Small with a patch size of 2 and image size of 32 for CIFAR-100 and Euro-SAT;
ViT-Small with a patch size of 16 and image size of 96 for STL-10; ViT-Small with a patch size of 16 and image size of 224 for Semi-Aves. To avoid the impact of the training configurations (e.g., batch size, learning rate, learning rate scheduler, etc.), we keep the settings described in~\cite{wang2022usb}. 
The pre-train model is from ~\cite{wang2022usb}, and we fine-tune the model on benchmarks using the proposed SSL algorithm. Our experiments are conducted on a server with an Intel i9 CPU and two RTX 3090 GPUs. We train our model with three random seeds and report the average error rate and range. The results of baseline methods are obtained from the latest commit of USB repository~\cite{github_usb_2023}.

\textbf{Hyper-parameters of \algo.} For all datasets, we set $\lambda_{u}=1 $, $\lambda_{mae}=0.01$ and $\lambda_{sdt}=0.5$ except Semi-Aves ($\lambda_{sdt}=0.95$) and TissueMNIST($\lambda_{sdt}=0.05$). The default masking rate for MAE is 0.3, except for Semi-Aves, which is 0.5. Pixel normalization on reconstruction target is only applied to experiments for Euro-SAT and TissueMNIST. The decoder's default depth (i.e., the number of Transformer blocks) is 4.

\begin{table*}[htbp]
\centering
\caption{Error rates on challenge datasets~\cite{wang2022usb}. We report averages and ranges under three random seeds. \textbf{Bold} indicates the lowest error rate and \underline{underline} indicates the second lowest error rate. Fully-supervised results of STL-10 and Semi-Aves are unknown because labels of unlabeled images are not available.}
\vspace{-3mm}
\label{tab:error_rate}
\scriptsize
\setlength{\tabcolsep}{3pt}
\resizebox{0.95\textwidth}{39mm}{
    \begin{tabular}{c|ll|ll|ll|ll|l}
\toprule
Method & \multicolumn{2}{c|}{CIFAR-100}                    & \multicolumn{2}{c|}{STL-10}     & \multicolumn{2}{c|}{Euro-SAT}  & \multicolumn{2}{c|}{TissueMNIST} & \multicolumn{1}{c}{Semi-Aves} \\
\hline
\# of labels & \makecell{200} & \makecell{400} & \makecell{40} & \makecell{100} & \makecell{20} & \makecell{40} & \makecell{80} & \makecell{400}   & \makecell{3959} \\
\midrule 
Pseudo Label                         & 33.99±0.95& 25.32±0.29& 19.14±1.3  & 10.77±0.6 & 25.46±1.36  & 15.7±2.12   & 56.92±4.54   & 50.86±1.79   & 40.35±0.3  \\
Mean Teacher                         & 35.47±0.4 & 26.03±0.3 & 18.67±1.69 & 24.19±10.15&26.83±1.46  & 15.85±1.66  & 62.06±3.43   & 55.12±2.53   & 38.55±0.21 \\
$\Pi$model                            & 36.06±0.15& 26.52±0.41& 42.76±15.94& 19.85±13.02&21.82±1.22  & 12.09±2.27  & 55.94±5.67   & \textbf{47.05±1.21}   & 39.47±0.15 \\
VAT                                  & 31.49±1.33& 21.34±0.5 & 18.45±1.47 & 10.69±0.51& 26.16±0.96  & 10.09±0.94  & 57.49±5.47   & 51.3±1.73    & 38.82±0.04 \\
MixMatch                            & 38.22±0.71& 26.72±0.72& 58.77±1.98 & 36.74±1.24& 24.85±4.85  & 17.28±2.67  & 55.53±1.51   & 49.64±2.28   & 37.25±0.08 \\
ReMixMatch                          & 22.21±2.21& 16.86±0.57& 13.08±3.34 & \underline{7.21±0.39} & \underline{5.05±1.05}   & 5.07±0.56   & 58.77±4.43   & 49.82±1.18   & \textbf{30.2±0.03}  \\
AdaMatch                           & 22.32±1.73& 16.66±0.62& 13.64±2.49 & 7.62±1.9  & 7.02±0.79   & 4.75±1.1    & 58.35±4.87   & 52.4±2.08    & 31.75±0.13 \\
UDA                                & 28.8±0.61 & 19.0±0.79 & 15.58±3.16 & 7.65±1.11& 9.83±2.15   & 6.22±1.36   & 55.56±2.63   & 52.1±1.84    & 31.85±0.11 \\
FixMatch                            & 29.6±0.9  & 19.56±0.52& 16.15±1.89 & 8.11±0.68& 13.44±3.53  & 5.91±2.02   & 55.37±4.5    & 51.24±1.56   & 31.9±0.06  \\
FlexMatch                           & 26.76±1.12& 18.24±0.36& 14.4±3.11  & 8.17±0.78& 5.17±0.57   & 5.58±0.81   & 58.36±3.8    & 51.89±3.21   & 32.48±0.15 \\
DASH                                & 30.61±0.98& 19.38±0.1 & 16.22±5.95 & 7.85±0.74& 11.19±0.9   & 6.96±0.87   & 56.98±2.93   & 51.97±1.55   & 32.38±0.16 \\
CRMatch                             & 25.7±1.75 & 18.03±0.2 & \underline{10.17±0.0}  & None     & 13.24±1.69  & 8.35±1.71   & \underline{54.33±2.83}   & 51.02±1.28   & 32.15±0.17 \\
CoMatch                             & 35.08±0.69& 25.35±0.5 & 15.12±1.88 & 9.56±1.35& 5.75±0.43   & 4.81±1.05   & 59.04±4.9    & 52.92±1.04   & 38.65±0.18 \\
SimMatch                            & 23.78±1.08& 17.06±0.78& 11.77±3.2  & 7.55±1.86& 7.66±0.6    & 5.27±0.89   & 60.88±4.31   & 52.93±1.56   & 33.85±0.08 \\
FreeMatch   & \underline{21.4±0.3}  & \underline{15.65±0.26}& 12.73±3.22 & 8.52±0.53& 6.5±0.78    & 5.78±0.51   & 58.24±3.08   & 52.19±1.35   & 32.85±0.31 \\
SoftMatch                          & 22.67±1.32& 16.84±0.66& 13.55±3.16 & 7.84±1.72& 5.75±0.62   & 5.9±1.42    & 57.98±3.66   & 51.73±2.84   & 31.8±0.22  \\
DefixMatch                          & 31.52±1.85& 21.12±1.74& 17.68±7.94 & 7.94±1.31& 14.71±6.52  & 3.72±0.79   & \textbf{54.07±6.19}   & \underline{48.95±1.14}   & 32.01±0.26 \\
\midrule 
\algo   &   \textbf{18.71±1.66}&           \textbf{14.87±0.88}&   \textbf{9.47±2.65}        &    \textbf{7.12±0.18}  &  \textbf{3.07±0.61}         &    \textbf{3.02±0.30}  &    54.94±1.10& 50.80±0.39    &    \underline{30.76±0.45}       \\
\midrule
Fine-Tune W/ Labeled Data                            & 35.88±0.36& 26.76±0.83& 19.0±2.9   & 10.87±0.49& 26.49±1.6   & 16.12±1.35  & 60.36±3.83   & 54.08±1.55   & 41.2±0.17 \\
Fully Supervised   & \multicolumn{2}{c|}{8.3±0.08}   & \multicolumn{2}{c|}{---} & \multicolumn{2}{c|}{0.94±0.03}     & \multicolumn{2}{c|}{28.96±0.17}   &  \multicolumn{1}{c}{---}  \\

\bottomrule
\end{tabular}
}
\vspace{-3mm}
\end{table*}

\subsection{Overall Result} 
We compare our method with existing SSL methods and report the top-1 classification error rates of CIFAR-100, STL-10, Euro-SAT, TissueMNIST, and Semi-Aves in Table~\ref{tab:error_rate}. \algo performs the best on the CIFAR-100, STL-10, and Euro-SAT datasets and achieves competitive results on the TissueMNIST and Semi-Aves datasets. It is notable that compared with strong baselines, \algo significantly decreases the error rate by 2.7 on CIFAR-100 and 2.0 on Euro-SAT with only two labels per class. On two settings (i.e., 4 and 10 labels per class) of the STL-10 dataset, we improve the state-of-the-art performance by 0.7\% and 0.1\%, respectively. This improvement is relatively small due to the inherently limited space for optimization. For the TissueMNIST dataset, the error rate of \algo is close to that of DefixMatch with only a 0.8 difference, while \algo is more stable with a smaller range of error rate, i.e., 1.1 since we achieve a stable feature extraction through a reconstruction target. For Semi-Ave, \algo reaches a competitive error rate of 30.76, only 0.56 away from that of ReMixMatch. The good performance of ReMixMatch on Semi-Ave may be because of reliable augmentation using augmentation anchoring ~\cite{berthelot2020remixmatch}. Our method could be applied to ReMixMatch to boost it.

\begin{figure}
    \centering
    \vspace{-2mm}
    \includegraphics[width=0.95\linewidth]{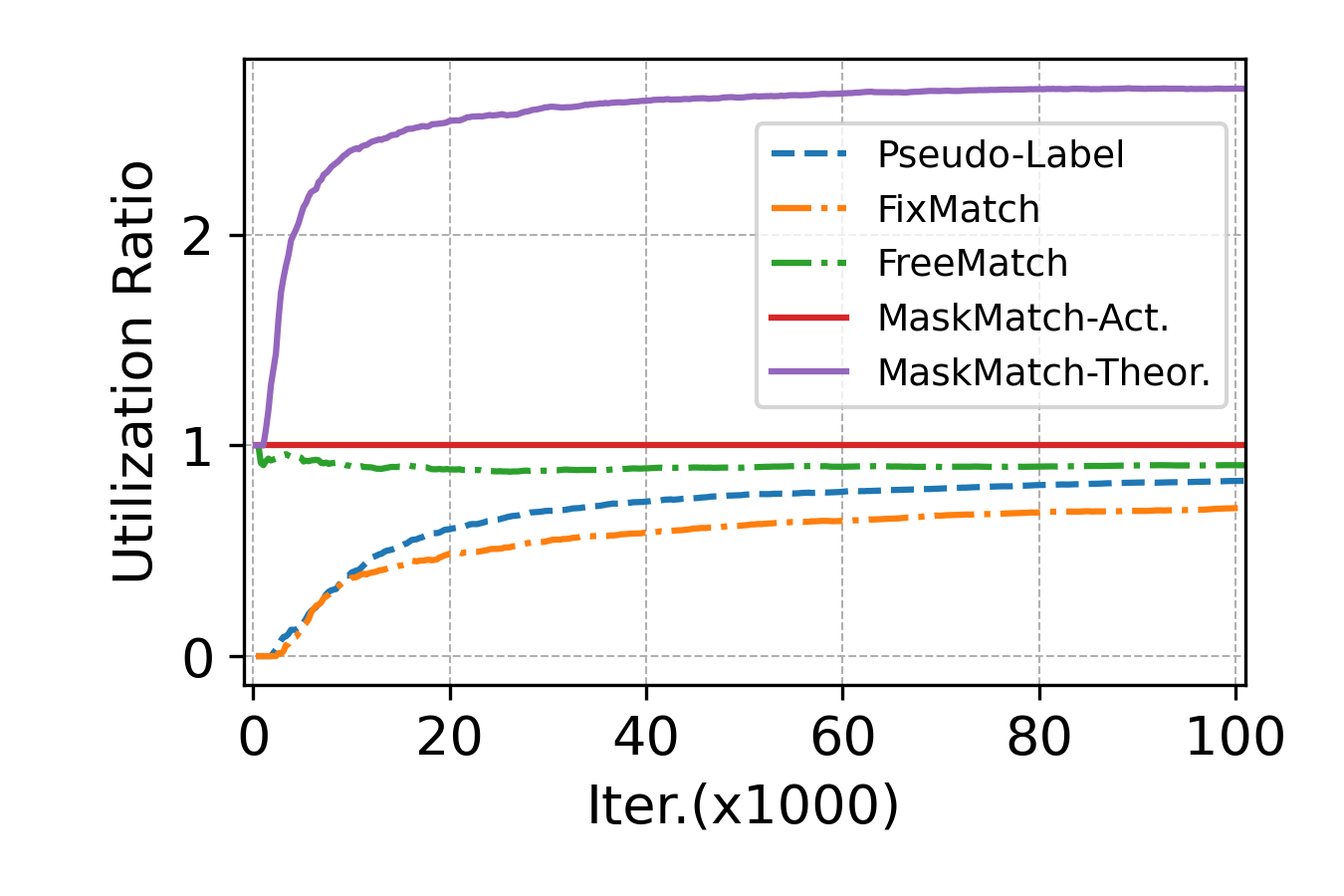}
    \vspace{-5mm}
    \caption{Data utilization on CIFAR-100 with 2 lables per class. MaskMatch-Act. and MaskMatch-Theor. represent the actual and theoretical data utilization, respectively.}
    \label{fig:total_utilization}
    \vspace{-2mm}
\end{figure}

\textbf{Data utilization.} We report the unlabeled image utilization to show that \algo can increase data utilization. The utilization is computed as $\frac{N_{train}}{N_{total}}$, where $N_{train}$ and $N_{total}$ represent the number of samples used in training and the total number of samples. Figure~\ref{fig:total_utilization} reports actual and theoretical utilization for \algo. The theoretical utilization counts one sample three times if this sample is involved in unsupervised loss, synthetic loss, and MAE loss. The actual utilization counts each sample once if this sample contributes to the total loss. From Figure~\ref{fig:total_utilization}, the actual utilization of \algo ($100\%$) is always higher than Pseudo-label~\cite{Pseudolabel}, FixMatch~\cite{fixmatch}, and FreeMatch~\cite{wang2023freematch} since MAE loss uses all images to improve the model's feature extraction. The theoretical utilization increases from $100\%$ and eventually stabilizes around $270\%$. In \algo, supervised loss and MAE loss dominate early, and then unsupervised loss and synthetic loss join training as the model's confidence improves.

\textbf{Computational cost.} The main concern of our method is computation overhead due to additional model computation on synthetic data and MAE encoder/decoder. We compare our training time with FreeMatch~\cite{wang2023freematch} on CIFAR-100 in Table~\ref{tab:training_time}. Overall, \algo increases the training time by 39.3\%. We also report the training time without MAE and SDT, respectively. The results show MAE requires more computation time than SDT due to the additional decoder.

\subsection{Ablation Study of Loss Terms}
According to Eq.~\ref{equ:total_loss}, \algo consists of multiple loss terms: supervised loss, unsupervised loss, MAE loss, and SDT loss. The benefits of supervised and unsupervised loss are well-studied in conventional SSLs~\cite{wang2023freematch, fixmatch, NEURIPS2019_mixMatch, berthelot2020remixmatch}. Here, our ablation study explores the benefits of MAE and SDT loss. We first introduce a baseline method that only uses supervised and unsupervised loss. Then, we explore the individual impact by adding each component (i.e., MAE and SDT) of \algo at a time and keeping other settings the same as the default. To show that SDT is better than MixUp augmentation, we also compare our SDT with MixUp augmentation (MixUp aug.)~\cite{NEURIPS2019_mixMatch} and its variant, Probabilistic Pseudo Mixup augmentation (ProbPseudo MixUp aug.)~\cite{cai2022semisupervised_psudomixup}. The ablation study is conducted on the Euro-SAT with 20 labels and CIFAR-100 with 200 labels (Table~\ref{tab:ablation_study}). 

\textbf{MAE loss.} Comparing baseline with MAE in Table~\ref{tab:ablation_study}, using MAE loss can significantly decrease error rate by 4.1 and 2.5 on the Euro-SAT and CIFAR-100 datasets, respectively. These results indicate that applying MAE constraints during SSL training can boost training performance. To dig deeply into the result, we defined optimization space (OS) as the error rate difference between the fully supervised in Table~\ref{tab:error_rate} and baseline in Table~\ref{tab:ablation_study}. Interestingly, MAE achieves greater improvement on Euro-SAT with a lower OS, 7.56, than on CIFAR-100 with a higher OS, 14.6. This is because the images in the Euro-SAT dataset are satellite images that are very different from the pre-training image dataset, ImageNet.
Thus, MAE in \algo can learn more from unlabeled images if they differ from pre-training images. 
This demonstrates the advantages of deploying \algo in real applications.

\textbf{Synthetic data training.} Comparing SDT with baseline in Table~\ref{tab:ablation_study}, SDT decreases the error rate by 3.8 and 1.0 on Euro-SAT and CIFAR-100, respectively. This result illustrates that SDT can improve training performance by using synthetic data. Comparing SDT with MixUp augmentations (e.g., MixUp aug. and ProbPseudo Mixup aug.), SDT decreases the error rate by 0.7 and 2.2, which is more than MixUp and ProbPseudo MixUp augmentation on Euro-SAT dataset. In addition, for the CIFAR-100 dataset, MixUp augmentation and ProbPseudo MixUp augmentation increase the error rate by 2.5 and 3.3. This could be due to MixUp augmentation leading to overfitting on noise~\cite{liu2023overtraining_mixup}. This result shows that SDT (i.e., using both original data and synthetic data) outperforms MixUp augmentation (i.e., using augmented data only).

\textbf{Updating initial threshold.} We also study the different initial class-specific thresholds, e.g., $\tau_0=1$ (default) and $\tau_0 =\frac{1}{\#class}$. Table~\ref{tab:ablation_study} shows that our initial threshold gets better performance. This is because introducing SDT requires accurate pseudo-labels.


\subsection{Configuration Study for MAE}
We also study the configurations for MAE, specifically the random masking ratio and the number of transformer blocks in the decoder, aiming to gain insights into the critical design decisions for MAE. Unless otherwise specified, the remaining configurations are kept consistent with the defaults outlined in Section~\ref{sec:setup}.

\begin{figure}
    \centering
    \includegraphics[width=0.95\linewidth]{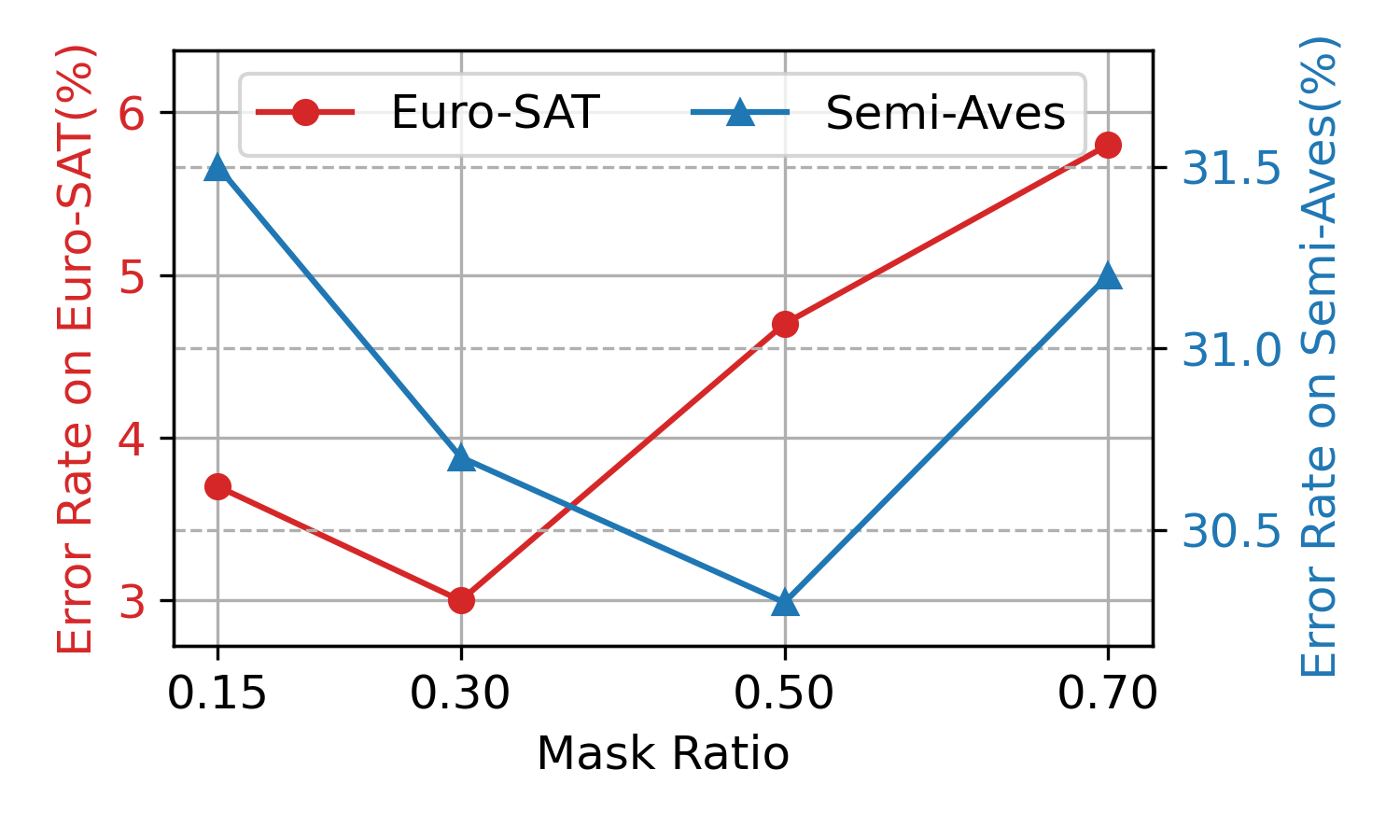}
    \vspace{-5mm}
    \caption{Error rate on Euro-SAT with 2 labels per class and Semi-Aves when varying the masking ratio. A suitable masking ratio is related to image size.}
    \label{fig:abs_maskrate}
\end{figure}

\textbf{Random masking ratio.}
We investigate the impact of masking ratio on model performance with two datasets: Euro-SAT with a small image size of 32 and Semi-Aves with a larger image size of 224. We vary the masking ratio from 0.15 to 0.70 and report the error rate in Figure~\ref{fig:abs_maskrate}. Our results indicate that the masking ratio plays a crucial role in the error rate and should be carefully balanced, avoiding either too high or too low. This finding can be explained by the gap between the reconstruction target and the recognition task. An excessively high masking ratio complicates the reconstruction process, diverting the model's focus from recognition tasks. Conversely, a very low masking ratio makes the reconstruction task overly simple, not adequately challenging the model's feature extraction abilities. 
From Figure~\ref{fig:abs_maskrate}, for smaller images (Euro-SAT of 32x32), the ideal masking ratio is around 0.3, whereas, for larger images (Semi-Aves of 224x224), a ratio of 0.5 is more suitable. Our hypothesis is that given the same masking ratios, smaller images are more prone to entirely mask out a locality, leading to the potential loss of critical information.

\textbf{Depth of decoder.} 
Due to the asymmetric architecture of MAE, the design of the decoder is not reliant on the in-training model (i.e., encoder), indicating the potential for selecting an appropriate decoder to enhance performance. In Table~\ref{tab:abl_depth}, we vary the decoder depth (i.e., the number of transformer blocks). Notably, a decoder with just a single block outperforms the existing methods on Euro-SAT with an error rate of 4.5. Further increasing the decoder depth demonstrates that a sufficiently deep decoder can enhance the SSL training, as shown in Table~\ref{tab:abl_depth}. This improvement can be attributed to the inherent difference between the image classification task and the reconstruction task. 
Although we use a reconstruction task to improve feature learning, the ultimate goal for our model is to excel in the image classification task. If the decoder is too shallow (e.g., single block) to reconstruct pixels, our in-training encoder will be specific to MAE loss and lose the generalization to the classification task. This results in the final layers of the encoder outputting specific representatives that are easily used for reconstruction but may not contribute significantly to the classification task. In contrast, a decoder with adequate depth can adapt to this difference and effectively specialize in reconstruction while allowing the encoder to focus on extracting more latent representations for image classification. By default, we use a 4-block decoder in our experiments.

\begin{table}
\centering
\caption{Training time on CIFAR-100 with 2 labels per class.}
\vspace{-2mm}
\label{tab:training_time}
\begin{tabular}{ccc}
\hline
\textbf{Algorithm} & \textbf{Minutes/epoch}  \\
\hline
FreeMatch & 2.7  \\
MaskMatch W/O MAE & 3.2 ($18.3 \% \uparrow$) \\
MaskMatch W/O SDT & 3.5 ($28.8 \% \uparrow$) \\
MaskMatch & 3.8 ($39.3 \% \uparrow$) \\
\hline
\end{tabular}
\vspace{-5mm}
\end{table}

\begin{table}
\centering
\caption{Ablation study. Baseline loss is $\mathcal{L}_{s} + \mathcal{L}_{u}$. We report the error rate under the same random seed.}
\vspace{-3mm}
\label{tab:ablation_study}
\begin{tabular}{ccc}
\hline
 \multirow{2}{*}{\textbf{Ablation}}& \multicolumn{2}{c}{\textbf{Error Rate\%}}  \\ \cline{2-3}
 & Euro-SAT & CIFAR-100 \\
\hline

Baseline & 8.5 & 22.9 \\
W/ MAE & 4.4(4.1$\downarrow$) & 20.4(2.5$\downarrow$) \\
W/ SDT  & 4.7(3.8$\downarrow$) & 21.9(1.0$\downarrow$) \\
\textbf{MaskMatch} & \textbf{3.0(5.5$\downarrow$)} & \textbf{18.1(4.8$\downarrow$)} \\
\hline
W/ MixUp aug. & 5.4(3.1$\downarrow$) & 25.3(2.5$\uparrow$) \\
W/ ProbPseudo Mixup aug. & 6.9(1.6$\downarrow$) & 26.2(3.3$\uparrow$) \\
\hline
MaskMatch ($\tau_0=\frac{1}{\#class}$) & 3.2(5.3$\downarrow$) & 19.3(3.6$\downarrow$) \\

\hline
\end{tabular}
\vspace{-2mm}
\end{table}

\begin{table}
    \centering
    \caption{Depth of Decoder. Deeper decoder decreases error rate.}
    \vspace{-3mm}
    \begin{tabular}{ccccc} \hline 
         \textbf{\# Transformer Block} & 1&  2&  4 & 8\\ \hline 
         \textbf{Error Rate}&  4.5&  4.1&  3.0& 3.0\\ 
         \textbf{Diff. to default}&  -1.5&  -1.1&  -- & 0.0\\ \hline
    \end{tabular}
    \vspace{-3mm}
    \label{tab:abl_depth}
\end{table}

\section{Conclusion and Future Work}

This paper addresses a common limitation observed in current semi-supervised learning (SSL) methods, which stems from the partial utilization of unlabeled data due to threshold-based filtering. We hypothesize that valuable training data are overlooked in this process, and hence impact model performance. To optimize data utilization, we propose \algo, a solution that integrates both Mask Autoencoder and synthetic data training. Empirical evaluations show that \algo outperforms existing SSL algorithms, achieving lower error rates on challenging datasets such as CIFAR-100, STL-10, and Euro-SAT. 

In future work, we plan to simplify \algo while preserving its performance. Our current design on unsupervised learning loss is based on the "Weak and Strong Augmentation" strategy~\cite{xie2020UDA}. We notice that random masking can be considered a form of strong augmentation. Our objective is then to utilize a random masking strategy to achieve comparable performance with reduced computational costs.  In addition, while this paper predominantly concentrates on the Vision Transformer (ViT) architecture, we are interested in extending our methodology to encompass other architectures, such as Convolutional Neural Networks (CNNs), via knowledge distillation. This will involve first training a ViT model using \algo and subsequently transferring the learned knowledge to a CNN.

{\small
\bibliographystyle{ieee_fullname}
\bibliography{egbib}
}

\end{document}